# From Shallow Waters to Mariana Trench: A Survey of Bio-inspired Underwater Soft Robots

Jie Wang[1], Peng Du[3], Yiyuan Zhang[1], Zhexin Xie[2]* and Cecilia Laschi[1]

1. Advanced Robotics Centre, Department of Mechanical Engineering, National University of Singapore; 117575, Singapore
2. Department of Mechanical and Energy Engineering, Southern University of Science and Technology; 518055 China
3. School of Marine Science and Technology, Northwestern Polytechnical University; 710072 China

E-mail: xiezx@sustech.edu.cn

**Abstract**

Sample Exploring the ocean environment holds profound significance in areas such as resource exploration and ecological protection. Underwater robots struggle with extreme water pressure and often cause noise and damage to the underwater ecosystem, whilebio-inspired soft robots draw inspiration from aquatic creatures to address these challenges. These bio-inspired approaches enable robots to withstand high water pressure, minimize drag, operate with efficient manipulation and sensing systems, and interact with the environment in an eco-friendly manner. Consequently, bio-inspired soft robots have emerged as a promising field for ocean exploration. This paper reviews recent advancements in underwater bio-inspired soft robots, analyses their design considerations when facing different desired functions, bio-inspirations, ambient pressure, temperature, light, and biodiversity , and finally explores the progression from bio-inspired principles to practical applications in the field and suggests potential directions for developing the next generation of underwater soft robots.

## 1. Introduction

The ocean encompasses over 70 percent of the Earth's surface. Given its vast ecosystem, the ocean plays a critical role in regulating the global climate and preserving numerous biodiversity and resources [1][2]. While over 95 percent of ocean depths remain unexplored, investigating these environments is crucial for resource development, ecological protection, and gaining insights into geological and physical processes [3].

Underwater robots are pivotal for ocean development, particularly in marine environmental sampling, monitoring, and protection [4][5]. As early as 1960s, the remotely operated vehicles (ROV) emerged and have since then experienced growth [6]. The Scorpio ROV, which gained widespread academic attention during a rescue operation for a Russian submarine, marked a peak in ROV popularity. Following this, autonomous underwater vehicles (AUV) became more popular and attracted the interest from numerous scholars worldwide due to their untethered autonomous positioning and navigation capabilities [7]. The 'Blue-Fin' AUV, known for its mature mechanical, electrical, and dynamic structure, along with its long cruising time and extensive operational range [8], was involved in the rescue operation for the world-shattering MH370 air crash. However, the ability of rigid bodies to bear the extreme water pressure at great depths poses a significant challenge limiting the further development of traditional AUVs [9][10]. Besides, the risk of damage caused by collisions with rigid structures during interactions with the environment cannot be overlooked from the perspective of marine ecological protection.

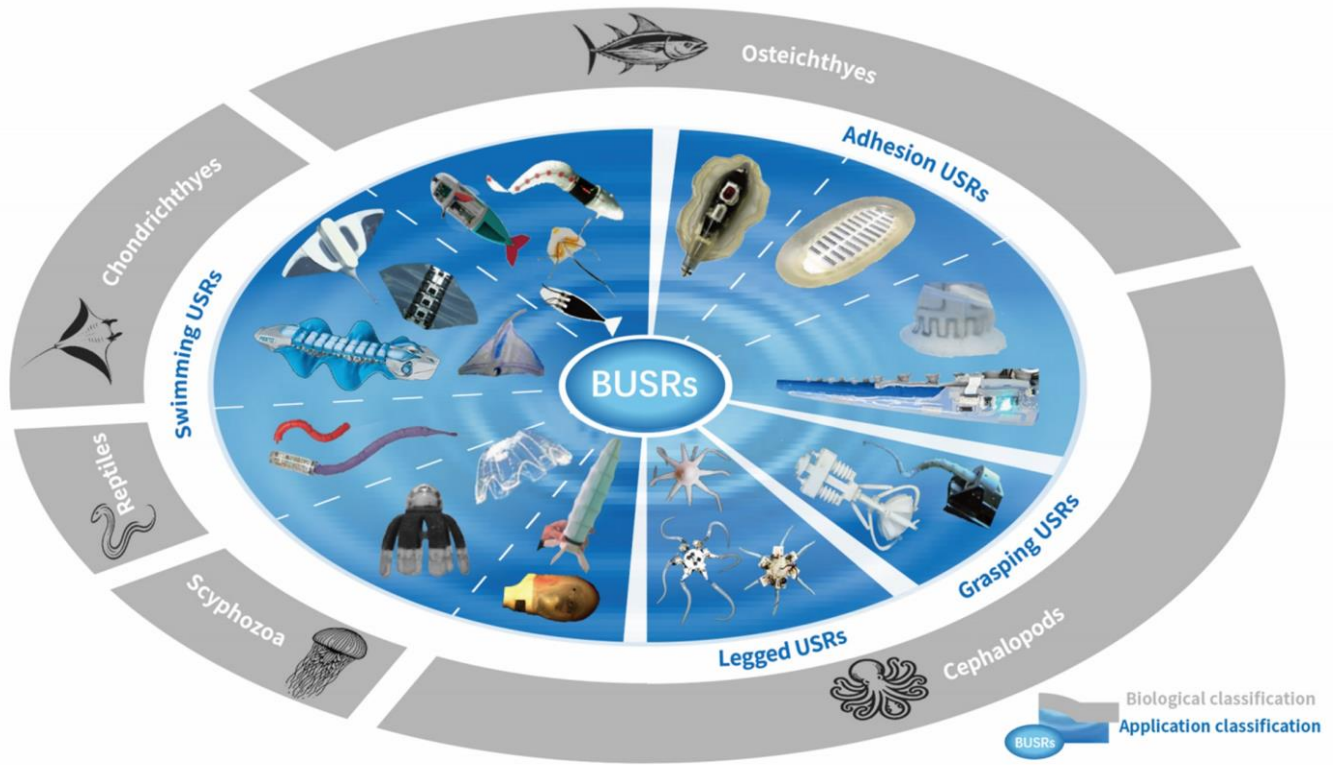

Fig. 1: Overview of Bio-inspired BUSRs

In contrast, underwater creatures generally possess soft bodies and have inherent advantages when navigating complex underwater environments. For instance, certain snailfish species (*Mariana snailfish, Ethereal snailfish*), discovered at depths of 8000 meters, possess hydrostatic skeletons that allow them to balance immense water pressure while maintaining the softness of their bodies for gentle interactions with environments [11]. Octopuses utilize embodied intelligence during the bending propagation of their arms to create efficient manipulation [12][13][14] significantly reducing water drag forces [14]. Soft robots draw inspiration from these aquatic creatures to enhance their underwater capabilities, not only in balancing water pressure and reducing drag but also in simplifying control strategies and interactive sensing of the environments. Additionally, the actuation of soft robots minimizes noise from cavitation effects and vibration of propellers [15], enabling their integration into ecological environments. Recent advancements in soft actuators and materials have facilitated the expansion of soft robots into underwater environments, utilizing actuation methods such as fluidic-driven, dielectric elastomer actuator (DEA)-driven, magnetic-driven, and thermal-driven systems [16]. Additionally, soft materials like silicone elastomers [17], hydrogels [18][19], DE [20], ionic polymer-metal composites (IPMCs) [21], and liquid metals [22] provide viable options in developing these underwater bio-inspired mechanical structures and their associated sensing systems. As a result, bio-inspired soft robots have been proposed throughout underwater environments, from shallow waters to the Mariana Trench.

Beyond existing considerable collections on recent advancements in bio-inspired underwater soft robots (BUSRs) [23][24][25], this paper explores how various considerations, including desired functions, bio-inspirations, ambient pressure, temperature, light, and biodiversity are closely related to the developments of BUSRs. To this end, this paper categorizes the BUSRs based on their reqired functions (swimming, legged locomotion, grasping, and adhesion) and the class of their biological models of inspiration (Osteichthyes, Chondrichthyes, FReptiles, Scyphozoa, and Cephalopods) (Fig. 1). Although certain classification schemes inevitably exhibit dimensional overlap, for instance, octopuses, are usually good bionic creature for both of grasping, crawling, and swimming robot design. We categorize them according to the most representative characteristics.

Furthermore, this paper explores how underwater environments, ranging from shallow to deep water, influence the motion performance, durability, and stability of soft robots. And in the end traces the evolution from bioinspired principle to field applications, addressing remaining challenges and proposing potential solution for the future development of the environment adaptive underwater robots.

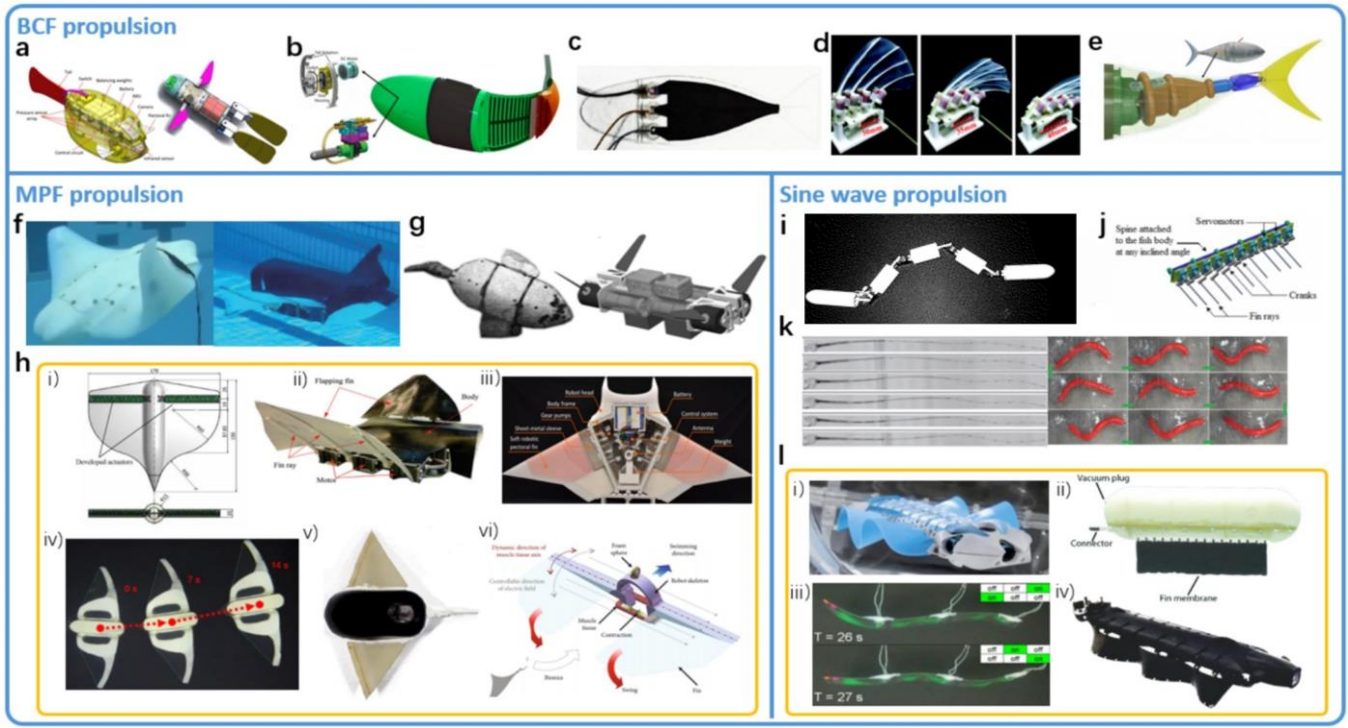

Fig. 2: Undulatory propulsion BUSRs. (**a**) **left** Biomimetic box puffer fish. Reproduced with permission from [29], copyright 2016, IEEE. (**a**) **right** Agile robotic double tailed fish. Reproduced with permission from [30], copyright 2018, IEEE. (**b**) Fluidic actuated fish with a cavity in the tail and gear pump driven hydraulic system or multiple solenoid valves controlled pneumatic system in the head. Reproduced with permission from [31], copyright 2014, Mary Ann Liebert, Inc. (**c**) DE-driven fish. Reproduced with permission under CC BY 4.0 license from [20]. (**d**) The bio-inspired soft-driven BCF BUSR with fordable dorsal fin. Reproduced with permission from [32], copyright 2014, Mary Ann Liebert, Inc. (**e**) Embedded hinge driven fish. Reproduced with permission from [33], copyright 2019, AAAS. (**f**) **left** 'MantaBot'. Reproduced with permission from [34], copyright 2011, IEEE. (**f**) **right** 'RoMan iii'. Reproduced with permission under CC BY 4.0 license from [35]. (**g**) **left** Pneumatic manta. Reproduced with permission from [36], copyright 2005, IEEE. (**g**) **right** Multi-fin flapping robot. Reproduced with permission from [37], copyright 2004, IEEE. (**h**) **i** Pneumatic actuated manta robot. Reproduced with permission from [38], copyright 2007, IEEE. (**h**) **ii** Motors actuated manta robot ('RoMan ii'). Reproduced with permission from [39], copyright 2007, Jilin University. (**h**). **iii** hydraulic actuated manta robot. Reproduced with permission under CC BY 4.0 license from [40]. (**h**). **iv** DE actuated manta robot. Reproduced with permission under STM PG from [41]. (**h**). **v** SMA actuated manta robot. Reproduced with permission from [42], copyright 2009, IEEE. (**h**). **vi** Living muscle-actuated manta robot. Reproduced with permission under CC BY 4.0 license from [43]. (**i**) Multiple hinges snake. Reproduced under STM PG from [44]. (**j**) Motors-driven system of fin-driven sine wave propulsion BUSRs. Reproduced with permission from [45], copyright 2005, IEEE. (**k**) **left** PFC driven snake. Reproduced with permission from [46], copyright 2024, IEEE. (**k**) **right** Pneumatic eel-like robot. Reproduced with permission from [47], copyright 2014, IEEE. (**l**). **i** 'BionicFinWave' developed by Festo, Germany. Reproduced with permission from Festo SE & Co. KG. (**l**). **ii** Undulating fin propulsion bio-inspired robot. Reproduced under STM PG from [48]. (**l**). **iii** DEA-driven transparent mute eel-like robot. Reproduced with permission from [49], copyright 2018, AAAS. (**l**). **iv** 'Velox' amphibious bionic robot made by Pliant Energy Systems, America. Reproduced with permission from Pliant Energy Systems Inc.

While some robots with rigid links and passive compliance have previously been described as "Soft Robotics" [26], this paper focuses on BUSRs that exhibit compliance and deformability in their materials and structures, in the interaction with the environment [27]. It follows the definition provided by the RoboSoft community: "Soft robot/device that can actively interact with the environment and can undergo 'large' deformations relying on inherent or structural compliance" [27]. Some exposed single-hinged structure underwater robots fall outside the scope of this paper, as they lack a compliant structure and cannot facilitate significant continuum deformation. Similarly, micro-robots inspired by microorganisms, such as single-celled organisms [28], are more likely to be used in medical applications rather in ocean exploration and development, therefore not included in this paper.

## 2. BUSRs for Swimming

The propulsion modes of swimming BUSRs can be divided into undulatory propulsion and jetting propulsion.

### 2.1 Undulatory Propulsion

The research on undulatory propulsion BUSRs began with an artificial fish named 'RoboTuna,' developed by Prof. Triantafyllou at the Massachusetts Institute of Technology in 1994. This tuna-inspired underwater robot features a streamlined body and a powerful flexible tail. To achieve high propulsion efficiency, the embedded servomotor-multi hinges drive system provides rhythmic continuous deformation of the tail [50]. Subsequent research led to an improved version of 'RoboTuna,' named 'VCUUV' has shown its stability, controllability, and high-speed performance [51]. Building on this success, numerous scholars have focused on further research regarding the 'flexible tail,' resulting in a wealth of related achievements over the past 30 years.

Notably, some bio-inspired fish are actuated by motors or servo systems, with their bodies connected to independent flexible plates through exposed hinges, such as the agile robotic double-tailed fish developed by the University of Science and Technology of China [30] and the multimodal swimming bionic box pufferfish developed by Peking University [29] (Fig. 2a). Although this kind of rigid-elastic hybrid robotic structure does not have the actively controllable deformed continuum body, It was widely adopted in the early related research of underwater bioinspired swimmer.

The propulsion principle of fish-inspired underwater robots involves swinging the tail fin along the vertical axis. In biological terms, this propulsion structure is referred to as "Body and/or Caudal Fin," abbreviated as BCF [52]. Three different waveforms are adopted by undulatory propulsion: the traveling-wave form, the mixture-wave form, and the standing-wave form [53]. The BCF propulsion BUSRs can be viewed as series-distributed bodies, comprising rigid heads and soft tails. The rigid components can contain electrical and electronic components and withstand high pressure by injecting insulating liquids, such as silicone oil [51]. The soft components provide mechanical oscillation for swimming. This sectional body design makes standing-wave actuators, and those driven by motors (with hinges completely embedded in the soft body).

The former typically employs fluid-driven systems or DE actuators. A gear pump or multiple solenoid valve systems facilitate fluid (gas or liquid) flow between two sides of the tail [31]. A cavity array in the tail (Fig. 2b) allows for significant deformation to achieve soft actuation. DE actuators (Fig. 2c) can deform in response to electrical stimulation, converting changes in the electric field into mechanical oscillations for soft actuation [20][55]. Compared to pneumatic actuators, hydraulic and DE actuators do not create significant pressure differences between the robot body and the surrounding environment [56]. This characteristic is essential for buoyancy adjustments, hydraulic and DE-driven systems that can avoid gas leakage issues during swimming [57].

Motor-driven robots typically feature a soft tail with embedded hinges, with drive motors installed either in the rigid heads or embedded within the soft tails. The tail acts as a hydrostatic body capable of withstanding high water pressure without requiring special treatments, such as the injection of insulating fluid. Driven by motors, the soft tail can achieve higher oscillation frequencies and larger continuum deformation. With the swing frequencies exceeding 10 Hz, surpassing those of most marine fish species [33], the bionic tuna produced by the University of Virginia is a first example of this type (Fig. 2e).

Interestingly, BCF fish adjust the retraction state of their fins based on their motion state while swimming. This phenomenon has been observed on largemouth bass (*Micropterus salmoides*) [58]. The area of the dorsal fins changes significantly when the fish is turns or accelerates [59][60]. To clarify the impact of this phenomenon on fish movements, a BCF BUSR with a foldable dorsal fin was developed (Fig. 2d), demonstrating that changes in dorsal fin area significantly affect fish acceleration [32].

Building on the research on BCF-based BUSRs, another type of fin-inspired underwater robot emerged, known as median and paired fin (MPF) propulsion BUSRs. This type of robots is inspired by cartilaginous fish, which possess bat-like fins [52]. Notable examples in this field, are bionic mantas and rays. The primary mode of MPF propulsion is through the flapping motion of ocean creatures. In fact, flapping propulsion is a specific case of traveling-wave propulsion. Through capture and calculation trajectory of the endpoint on the manta thoracic fins, the continuous motion curve of the pectoral fins exhibits a distinct sine characteristic [34].

The reason why the flapping propulsion marine animals do not display distinct sine waves on their bodies is that the size of their pectoral fins is smaller than the wavelength [61]. MPF BUSRs feature a wing-body integration structure, consisting of a central columnar rigid body and a pair of large, flat, soft wings integrated on both sides [62]. This design minimizes water resistance during swimming by optimizing the airfoil-like cross-section area and surface turbulent flow layer. Turning can be easily accomplished through asymmetric fin motion [62][63]. Compared to BCF propulsion, MPF propulsion offers higher cruise efficiency because it can glide using a large median and paired fin [34]. This capability suggests that flapping BUSRs may operate effectively in larger water bodies due to their longer endurance.

The development of MPF BUSRs dates to around 2010 [64]. The initial research on manta-inspired robots were conducted by a team that included Prof. Alexander J. Smits from Princeton University, Prof. H. Bart-Smith from the University of Virginia, and Prof. Frank Fish from West Chester University [65]. A notable example is the 'MantaBot' (Fig. 2f, left), a series of soft biomimetic manta robots with high propulsive efficiency [66], which continues to be updated to the present. Around the same time, Nanyang Technological University produced their first vision soft manta robots, named 'Ro-Man,' and subsequently published several improved versions (Fig. 2f, right). The National University of Singapore later developed a 50-kilogram hydraulic manta robot. Some early rigid manta-inspired robots, which connected independent 50-kilogram flat fins to the body with exposed rigid hinges, are not classified as soft robots in this paper (Fig. 2g).

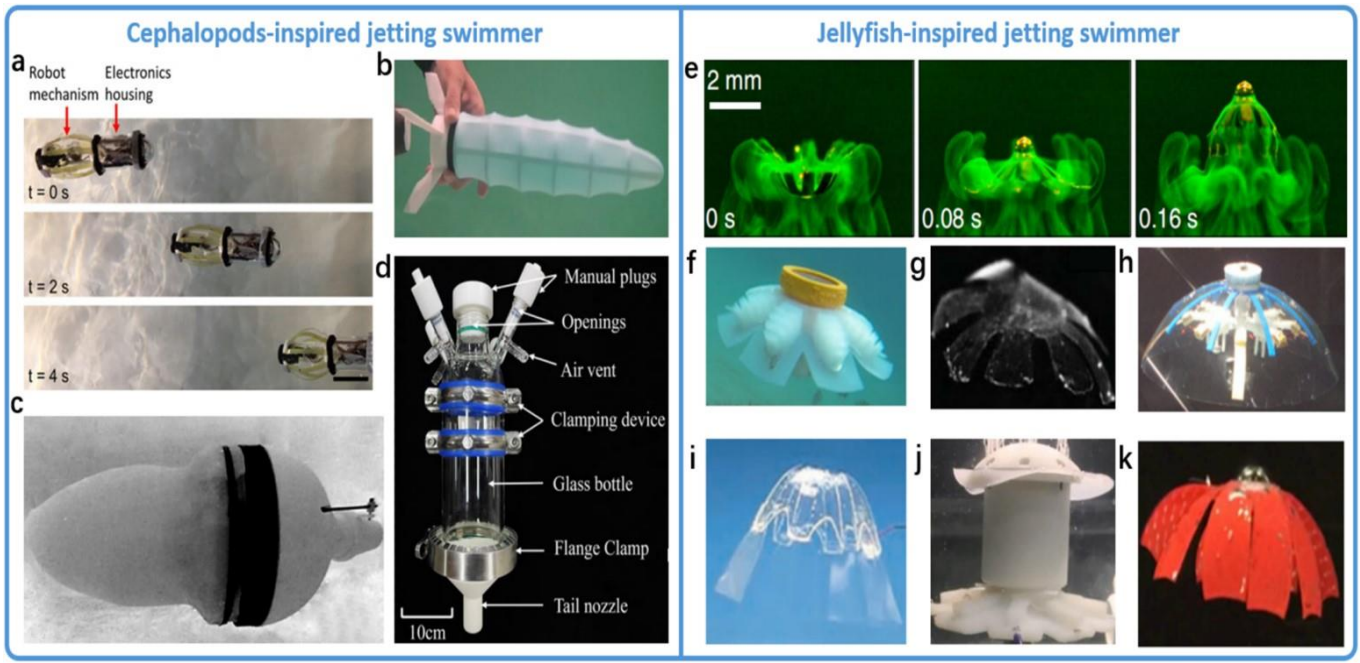

Fig. 3: Jetting propulsion BUSRs. (**a**) Motor and slip-gear driven jetting swimmer. Reproduced under STM PG from [67]. (**b**) Elastic membrane-driven single-launched jetting swimmer. Reproduced under STM PG from [68]. (**c**) Pulsed-jet propulsion swimmer with vector steerable nozzle. Reproduced with permission from [69], copyright 2019, IEEE. (**d**) High-pressure gas-driven jetting swimmer. Reproduced with permission under CC BY 4.0 license from [70]. (**e**) Magnetic composite elastomer-driven jellyfish BUSR. Reproduced with permission under CC BY 4.0 license from [71]. (**f**) Hydraulic-driven jellyfish BUSR. Reproduced under STM PG from [72]. (**g**) Muscle tissue-driven jellyfish BUSR. Reproduced under STM PG from [73]. (**h**) IPMC driven jellyfish BUSR. Reproduced under STM PG from [74]. (**i**) Transparent jellyfish BUSR. Reproduced with permission from [75], copyright 2023, Mary Ann Liebert, Inc. (**j**) Pneumatic driven jellyfish BUSR. Reproduced with permission under CC BY 4.0 license from [76]. (**k**) DE driven jellyfish BUSR. Reproduced with permission under CC BY 4.0 license from [77].

The primary technology of MPF BUSRs involves driving driving the median paired fin to flap. To date, MPF BUSRs have employed various actuation methods (Fig. 2f-h), including motors [35], hydraulic actuation [36], pneumatic actuation [34], shape memory alloys (SMA) [42], living muscle tissues [43], and DE [78]. Benefiting from the pressure-resistant properties of DE and silicone matrices, soft robots have been successfully applied in the Mariana Trench [41]. In addition to MPF propulsion BUSRs, another type of soft robot utilizing traveling-wave propulsion is the sine wave propulsion ones. This category of BUSRs can be divided into two types: eels [79][80], snakes [46] inspired whole-body driven BUSRs, and the knife-fishes [81] inspired fin-driven BUSRs.

Eels and sea snake robots feature bodies that can be continuously deformed to produce sine waves. The early development of these robots involved multiple hinges arranged in series (Fig. 2i). The control and locomotion performance of this traditional driving method is limited by the size of servo motors and hinges. The adoption of soft actuators raised the oscillation frequency and propulsion efficiency. A piezoelectric fiber composite (PFC) driven bio-inspired snake (Fig. 2k, left) can achieve oscillation frequencies exceeding 40 Hz and propulsion velocities of nearly 0.6 m/s propulsion velocity, beyond the real sea snakes [46].

Fin-driven sine wave propulsion BUSRs exhibit greater adaptability to varying water flow conditions and can maintain static suspension due to their more flexible and controllable fins [81][82]. The artificial narrow and long fin is often actuated by a mechanical structure (multi-link and cranks) with servo motors (Fig. 2j), such as the Nanyang Knifefish (NKF) [45], a series of sine wave propulsion BUSRs that utilize a multi-link structure [83]. There are two fin-body combination modes: vertical connection and symmetrical horizontal connection. The phase differences between fin rays (the ribs perpendicular to the wave propagation direction) result in variable wavelengths and amplitudes generated by a timing control system. Various mechanisms have been adopted in many classical prototypes (Fig. 2l).

*2.2 Jetting Propulsion*

Jetting propulsion is primarily utilized by cephalopods and jellyfish. Octopuses and squids are representatives of cephalopods, capable of achieving high escape velocities when threatened. An octopus can accelerate to impressive speeds by expelling water from its cavity [84][85]. This rapid and powerful process relies on the mantle muscle of the octopus [86]. Researchers have developed bio-inspired jetting swimmers based on this characteristic. Different jetting propulsion methods exist, such as a single-launched swimmer (Fig. 3b) driven by the elastic force of a cover film, achieving speeds exceeding ten times its body length per second (BL/S) [68]. Additionally, the cavity-membrane-based water-jet (CM-jet) structure (Fig. 3d) allows the high-pressure gas to push the liquid from the chamber, generating high-speed jets. This driven mode was inspired by the flying squid [70]. However, this single-time propulsion method cannot provide sustained and stable thrust over extended periods. Consequently, motor-driven jetting swimmers with slip-gear transmission mechanisms (Fig. 3a) have been developed to achieve continuous propulsion [67]. In nature, some octopuses and squid adopt pulsed-jet propulsion. Bio-inspired pulsed-jetting swimmers are driven by cables and crank rotation, proving more efficient due to vortex ring formation [87][88][89]. Furthermore, vector steerable nozzles enhance the cornering capabilities of jetting swimmers (Fig. 3c).

Jetting propulsion is also a primary swimming method for jellyfish [90]. Jellyfish (medusae) are cnidarian creatures with bell-shaped bodies. While some jellyfish swim by rowing their bell margins, most generate thrust through the overall contraction of their bells. Early prototypes of bio-inspired jellyfish were driven by motors with rigid links, providing mechanical output at fixed frequencies. Subsequently, more products have adopted soft and smart actuators, such as SMA, fluid-driven systems, DE, muscle tissue, and IPMC (ionic polymer–metal composite).

In 2009, research on soft jellyfish robots driven by SMA began [91]. Then, the IPMC-driven jellyfish BUSRs and their upgrades were developed by Prof. Joseph S. Najem and his team (Fig. 3h). IPMC is a type of artificial muscle whose underwater performance, particularly for propulsion, was detailed in research conducted in 2007 [92]. IPMC addressed the high power consumption issues with SMA, achieving much higher swimming velocities [74][93]. Moreover, living muscle tissue was adopted on jellyfish BUSRs (Fig. 3g). Cultured myocardial tissue can respond to electrical fields to generate thrust, propelling underwater robots [73].

Moreover, fluid actuators were adopted on jellyfish BUSRs, including hydraulic-driven (Fig. 3f) and pneumatic-driven (Fig. 3j) systems. While pneumatic actuators require an external air source, limiting their operational range, they provide high-pressure gas, enabling jellyfish faster and carry heavier loads. Hydraulic actuators can be designed as untethered systems, allowing BUSRs to swim freely. Thus, the hydraulically driven jellyfish have successfully reached depths of over 300 meters [72]. Additionally, DE-driven jellyfish have emerged (Fig. 3k), with some even being transparent in water, reducing their negative impact on natural biological systems (Fig. 3i).

| BCF BUSRs | Body Length/Weight (mm/g) | Speed (BL/s) | Corresponding Frequency (Hz) |
|---|---|---|---|
| RoboTuna (1996) [46] | 2400/173000 | 0.65 | <1.00 |
| VCUUA (2002) [47] | -/- | 0.61 | <1.00 |
| Pneumatic Tuna [31] | 339/- | 0.44 | 1.67 |
| DEA Driven Fish [20] | 150/4.4 | 0.25 | 0.75 |
| DEA Driven Fish [51] | 175/- | 0.14 | 3.00 |
| Morphing Fin fish [28] | 588/2790 | 0.64 | 1.50 |
| Tunabot [29] | 255/306 | 4.00 | 15.00 |
| SoFi [53] | 470/1600 | 0.51 | 1.40 |
| **MPF BUSRs** | **Body Length/Weight (mm/g)** | **Speed (BL/s)** | **Corresponding Frequency (Hz)** |
| MantaBot [30][59] | 430/- | 0.81 | 1.82 |
| RoMan III [35] | 370/5000 | 0.81 | 1.50 |
| Pneumatic Manta [38] | 150/- | 0.67 | - |
| Hydraulic Manta [40] | 850/48600 | 0.41 | 0.25 |
| DEA Driven Manta [41] | 93/- | 1.45 | 5.00 |
| SMA Driven Manta [42] | 220/354 | 0.26 | 1.20 |
| Living Muscle Driven Manta [43] | 13/- | 0.006 | 2.00 |
| **Sine Wave Driven BUSRs** | **Body Length/Weight (mm/g)** | **Speed (BL/s)** | **Corresponding Frequency (Hz)** |
| PCF Driven Snake [46] | 370/8.3 | 0.09 | 1.70 |
| Pneumatic Snake [47] | 530/- | 0.20 | 1.25 |
| Motors Driven Knife Fish [48] | 459/- | 0.65 | 6.00 |
| DEA Driven Eel [49] | 220/- | 0.01 | 0.33 |
| **Jetting Propulsion BUSRs** | **Body Length/Weight (mm/g)** | **Speed (BL/s)** | **Corresponding Frequency (Hz)** |
| Motor Driven Octopus [67] | 341/- | 0.54 | 1.12 |
| Elastic Membrane Driven Squid [68] | 360/1030 | 10.00 | - |
| Vortex Propulsion Octopus [69] | 160/333.5 | 0.25 | 1.50 |
| **Jellyfish Inspired BUSRs** | **Body Length/Weight (mm/g)** | **Speed (mm/s)** | **Corresponding Frequency (Hz)** |
| Hydraulic Jellyfish [72] | 210/- | - | 0.80 |
| IPMC Driven Jellyfish [74] | -/- | 1.50 | 0.50 |
| Pneumatic Jellyfish [76] | -/- | 160 | 0.80 |
| DEA Driven Jellyfish [77] | -/230 | 3.20 | 0.20 |

Table 1: Parameter statistics on swimming BUSRs.

It is well known that electromagnetic waves have limited travel capabilities in water, being suitable for short distances. This limitation has led many underwater vehicles to adopt wired control schemes. Sonar has the highest communication efficiency and stability underwater. However, it is not applicable to BUSRs, because of its rigid structures and high weight. For remotely operated BUSRs, magnetic fields may offer a promising solution. Consequently, a remotely operated magnetic composite elastomer-driven jellyfish was developed (Fig. 3e), capable of executing five different swimming modes in a remote magnetic field [71].

Currently, monolithic 3D printing technology has been employed for the fabrication of jellyfish-inspired BUSRs. This additive manufacturing method has significantly high efficiency while minimizing assembly errors. It improves the static stability of BUSRs due to the uniformity and symmetry of mass distribution [94].

For swimming BUSRs, the primary focus is on motion performance. The size, weight, and velocity parameters of several classical robots utilizing diverse actuation methods have been compiled (Table 1).

*2.3 Characterization of Various propulsion methods*

BCF and MPF swimmers employing Undulatory propulsion and those utilizing jetting or pulsed propulsion exhibit distinct hydrodynamic characteristics. Moreover, the low-temperature conditions of the deep-sea increase seawater viscosity, leading to a corresponding decrease in the Reynolds number [95][56]. Consequently, the adaptability and locomotion performance of different propulsion methods at varying ocean depths, including factors such as stability and propulsive efficiency, also differ significantly.

BCF and MPF propulsion represent the two most widely adopted locomotion modes in fish as well as in bionic soft swimmers. BCF propulsion typically relies on a relatively large driving force to actuate the caudal fin, which strongly interacts with the surrounding fluid [57][58]. This interaction generates alternating large-scale vortex streets that are subsequently converted into forward thrust [99]. Throughout this process, the Reynolds number remains at a high level [100]. A stronger driving force leads to a higher swimming speed, while lower fluid viscosity further enhances propulsion efficiency under BCF modes. In contrast, MPF propulsion is driven by the alternating oscillations of median or paired fins, usually at relatively low frequencies. This mechanism produces continuous local vortices around or at the tips of the fins, which provide thrust or lift [101][102]. Unlike BCF propulsion, which is dominated by inertial forces, MPF propulsion is governed primarily by viscous resistance. Consequently, under low-speed, high-viscosity, and low-Reynolds-number conditions, MPF exhibits superior swimming performance.

For bionic soft swimmers, the large torques generated by the caudal fin during BCF propulsion induce coupled yaw and roll moments [103], posing significant challenges for precise attitude control. By contrast, MPF-driven robots often feature a relatively flat body morphology; in some cases, such as manta-inspired designs, fin–body fusion reduces cross-sectional resistance and improves stability. Moreover, bionic soft robots are usually powered by artificial muscles, whose actuation speed and force output are considerably inferior to the muscular systems of real fish. Although the efficiency of BCF propulsion can even doubled through the variable stiffness tail fin technology [104], high output power is still necessary. As a result, achieving rapid swimming at high Reynolds numbers, and thereby attaining high propulsion efficiency remains difficult without traditional motors. Conversely, operating at lower swimming speeds can reduce gesture control complexity. Collectively, these factors render MPF propulsion modes more suitable for deep-water environments.

Jetting propulsion, on the other hand, relies on the direct thrust of the expelled jet flow as well as the reaction force from the vortex rings generated during jetting [105][106]. In emergency situations, squids primarily employ high-speed jets for rapid escape, whereas under routine conditions they switch to MPF propulsion. High-speed jets are better suited to low-viscosity fluids and require large instantaneous driving forces to eject high-pressure fluid from the mantle cavity. However, this propulsion strategy is unsuitable for BUSRs that prioritize stability and efficiency in deep-water applications. Certain jellyfish species utilize pulsed jets to generate vortex rings that provide thrust. Vortex rings produced at lower jetting speeds tend to be more coherent, thereby minimizing energy loss associated with vortex fragmentation. Additionally, the use of an embedded series oscillator enables more efficient utilization of eddy currents, while the resonance effect improves propulsion efficiency, thereby reducing the cost of transport (COT) of the squid inspired jetting robot to as low as 0.09, lower than real jellyfish [107]. Although generating low-speed jets does not demand powerful actuators or fast-response controllers and thus provides a theoretical foundation for adapting bionic soft robots to deep-sea exploration, the stability and controllability become huge challenges. Specifically, most jellyfish in nature drift passively with ocean currents as well. Thus, vectoring nozzles are required to solve the problem [108], which cause kinds of design and engineering challenges.

## 3. BUSRs for Legged Locomotion

While legged locomotion is inefficient in water, due to the water drag acting on rigid legs, some marine species show efficient locomotion, using soft or compliant limbs. The octopus soft arm is a versatile limb used for grasping and manipulation, as well as for walking [109]. Early research on the biological movement of octopuses highlights crawling as a highly efficient yet relatively simple motion, especially when compared to more complex actions like entanglement and grasping.

The walking ability of octopuses primarily relies on the push force generated by their arms. They first contract the proximal segment of their back arms, then fix the arms to the base using suckers, and finally elongate the proximal segment to complete a push cycle (Fig. 4a). To replicate this crawling motion in bio-inspired robots, early research concentrated on single arm's design and control. The soft arms are typically made of silicone and driven by embedded cables along the longitudinal direction (Fig. 4b). To enhance movement in all directions, multiple crawling arms have been developed, arranged in a circumferential distribution (Fig. 4b and Fig. 4f) [110]. To increase the efficiency of push and/or pull action cycles, a crank connecting rod structure driven by motors have been adopted, providing.4 stable mechanical output through cable reeling and releasing [111][112]. Additionally, fluidic actuators have proven effective for driving soft arms. Pneumatic-driven octopus arms can perform twisting and turning motions through the cooperation between multiple bending soft arms (Fig. 4c). By embedding a compilation network structure within the silicone arms, the radial pressure-bearing capacity of the chambers is reinforced, reducing the ballooning effect [113]. This embedded structure also allows for more internal space to install cables while maintaining bending and retraction

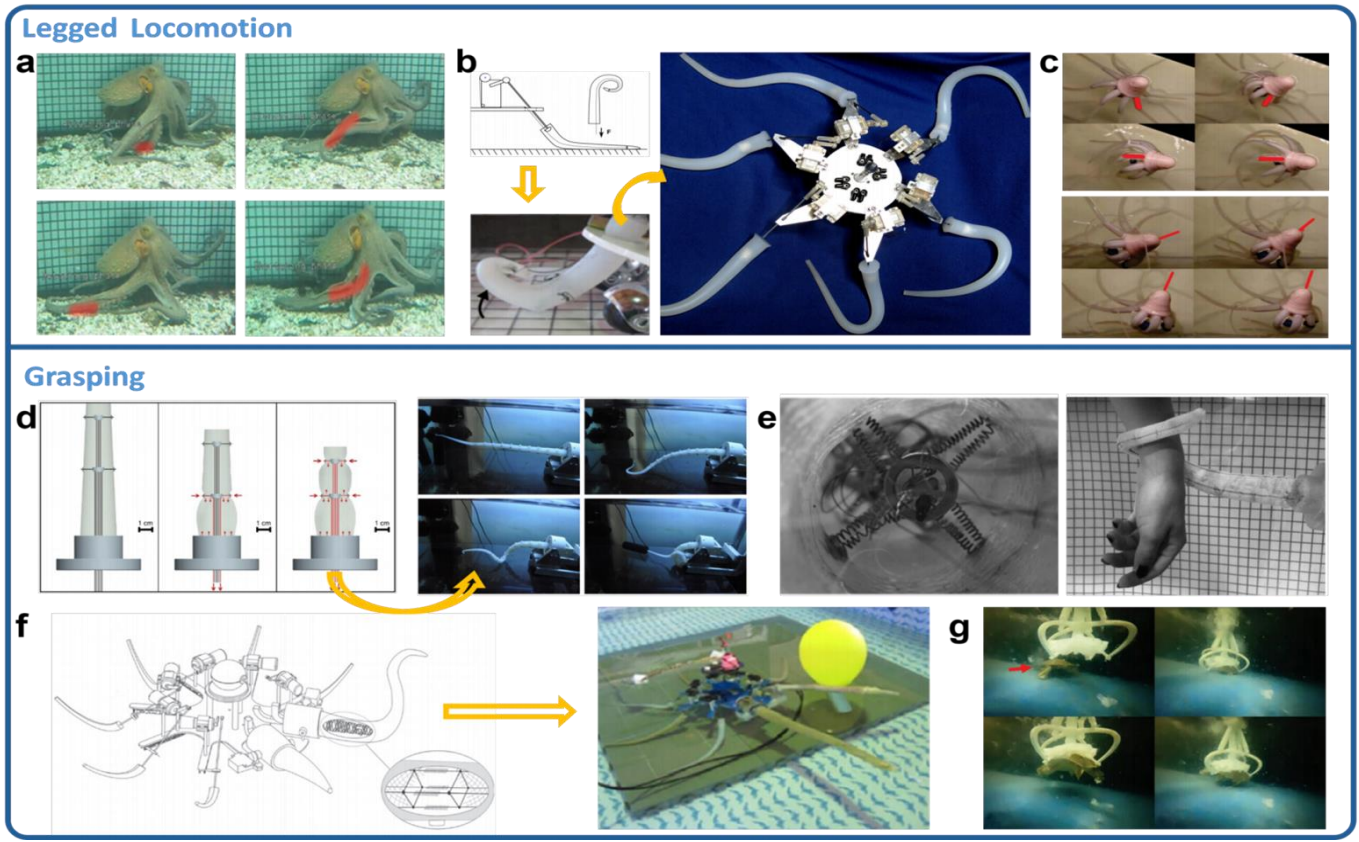

Fig. 4: Legged and Grasping BUSRs. **(a)** The crawling process of living octopus. Reproduced with permission under CC BY 4.0 license from [112]. **(b)** 6 arms octopus-inspired BUSR based on a Motor-crank structure, developed by Scuola Superiore Sant'Anna, Italy. Reproduced under STM PG from [110], with permission from [88], copyright 2013 MTS and [111], copyright 2011, IEEE. **(c)** Pneumatic crawling octopus' robot int twisting and turning. Reproduced with permission from [113], copyright 2011, IEEE. **(d)** Transverse and longitudinal cables driven soft arms with grasp and entangling ability. Reproduced with permission from [114], copyright 2011, Elsevier. **(e)** Transverse SMA and longitudinal cables driven soft arms with grasp and entangling ability. Reproduced under STM PG from [115]. **(f)** The 8-arm octopus-inspired robot with both crawling and grasp ability and embedded touching sensors. Reproduced with permission under CC BY 4.0 license from [112]. **(g)** Glowing sucker octopus-inspired gripper BUSR. Reproduced with permission under CC BY 4.0 license from [116].

capabilities [112].

## 4. Grasping BUSRs

Based on the biological principles of the octopus muscular hydrostat, transverse and longitudinal cables were integrated in a silicone arm, to reproduce the coupled longitudinal and transverse contractions during reaching movements (Fig. 4d). This configuration enables controllable bending through the coordination of longitudinal and transverse drive cables, enhancing stiffness through antagonistic effects [114].

Following this, soft octopus arms with longitudinal and transverse SMA springs were developed (Fig. 4e). The coupling effect of the driving force from longitudinal and transverse cables was decoupled, as the transverse-driven cables were replaced by a reticular formation consisting of SMA, allowing for independent actuation from the longitudinal cables [115].

Furthermore, using multiple arms for locomotion (for walking) and manipulation (for grasping) provides an effective approach to a more complete octopus robot (Fig. 4f) capable of performing complex tasks. In addition, the octopus-inspired robot gains environmental awareness through touch sensors [112], by embedding deformation-sensitive materials in the silicon arms, which can convert expansion and retraction into electrical resistance.

Researchers have not only developed soft manipulators for crawling and grasping by simulating biological octopuses but have also proposed novel soft gripper designs (Fig. 4g) inspired by bioluminescent octopuses (*Stauroteuthis syrtensis*). Bioluminescent octopuses possess a different body structure compared to other common cephalopod subspecies. They have a soft thin film between their arms with suckers, forming a large umbrella-shaped structure on the oral side [117]. This unique structure enables these octopuses, such as the vampire squid, to grasp effectively [118]. Drawing inspiration from this feature, research on soft robotic grippers has gained traction, demonstrating high performance in grasping irregular, scattered, and out-of-reach objects [116].

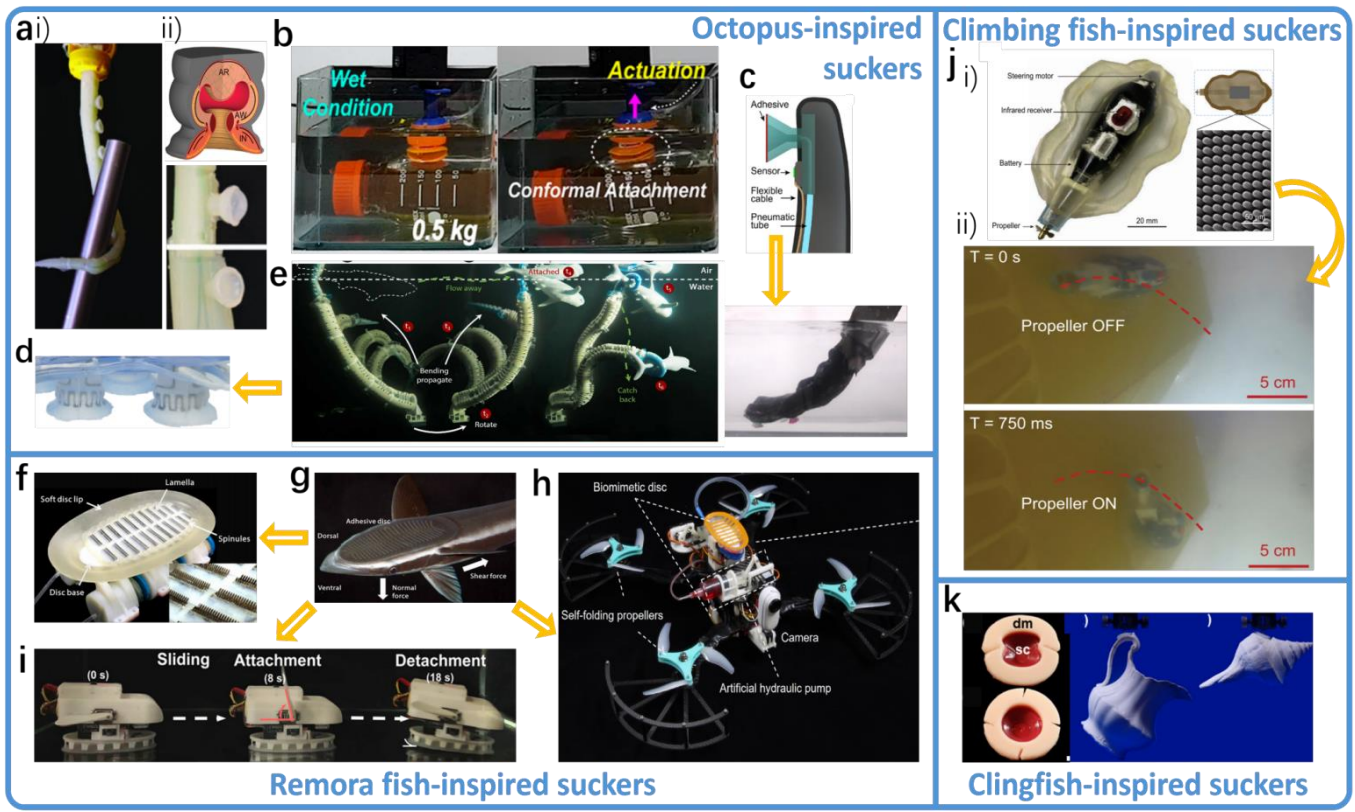

Fig. 5: Adhesion BUSRs. **(a). i)** Bioinspired octopus' soft arm with suckers for enhancing the grasping capacity. Reproduced with permission under CC BY 4.0 license from [122]. **(a). ii** Morphological structure of octopus sucker [123], and silicone bionic sucker with inflation and suction capacities. Reproduced with permission under CC BY 4.0 license from [122]. **(b)** Octopus-inspired sucker with carbon nanotube strain sensors. Reproduced with permission from [124], copyright 2018, ACS. **(c)** Octopus-inspired sucker with optical approach sensors. Reproduced with permission under CC BY 4.0 license from [125]. **(d)** Soft octopus-inspired suckers with embedded liquid metal sensors. **(e)** Octopus-inspired soft arm with suckers and the capacity to reach motion. Reproduced with permission from [126], copyright 2023, AAAS. **(f)** Remora-inspired pneumatic suction cup. Reproduced with permission from [127], copyright 2017, AAAS. **(g)** The biology structure of remora fish's sucker. Reproduced with permission from [127], copyright 2017, AAAS. **(h)** The aerial-aquatic robot with Remora-inspired hydraulic sucker. Reproduced with permission from [128], copyright 2022, AAAS. **(i)** Servo motor driven untethered Remora-inspired adhesion BUSR. Reproduced under STM PG from [129]. **(j). i)** Rock climbing fish-inspired multiple suction cups BUSR. Reproduced with permission under CC BY 4.0 license from [130]. **(j). ii)** The performance of dynamic adhesion of rock climbing fish-inspired BUSR. **(k)** Clingfish-inspired passive suction cups and the wet adhesive effect. Reproduced under STM PG from [131].

## 5. Adhesion BUSRs

In nature, some creatures adhere to other objects through various adhesion mechanisms, such as land creatures like caterpillars, tree frogs, and geckos. These mechanisms primarily rely on capillary action, mechanical interlocking, and Van der Waals force [119][120][121]. However, in underwater environments, creatures have evolved different adhesion mechanisms, such as suction. Due to the differences in environmental media, dry adhesion is not easily applicable in underwater settings [133].

Octopus-inspired soft suckers are among the most popular adhesion mechanisms in BUSRs. The biological structure of an octopus suction cup is a complex system consisting of the acetabulum, constricted orifice, and infundibulum [123]. It generates suction by creating a negative pressure chamber through seawater expulsion.

Additionally, numerous tactile and biochemical sensing nerves are distributed around the sucker to detect the physical and chemical properties of objects [134]. Based on biological principles and morphology, a pneumatic octopus-inspired soft sucker was developed with an optical approach sensor (Fig. 5c). This design can detect the distance to an object and automatically execute adhesion [125]. Recent research has also designed octopus-inspired soft suckers with liquid metal sensors embedded in an octopus-inspired soft arm (Fig. 5d). These sensors can detect adhesion and detachment status, providing feedback to a finger cuff that synchronously controls the soft arm (with suckers) to perform bending, adhesion and reaching motions (Fig. 5e). This work enhances the functionality of the biomimetic octopus arm and advances human-machine interaction in the human-machine interaction of soft robots. Notably, living octopuses utilize a soft dome-shaped protrusion to control adhesion and detachment [123]. This dome, located at the center of the sucker, can expand and contract to adjust the suction cup's pressure [135]. Research has developed soft suckers with protrusions and carbon nanotube sensors distributed on the surfaces of suction cups (Fig. 5b), enabling the calculation of the weight of adhering objects [124]. Additionally, studies have explored octopus-inspired soft arms with suckers that possess grasping and winding capabilities [122].

| Adhesion BUSRs | Sucker Size (mm) | Adhesive Force (N) | Adsorption Strength (Kpa) |
|---|---|---|---|
| Octopus Inspired Sucker [122] | Φ 32 | 45 | 56.25 |
| Remora Inspired Sucker [127] | 127*72 | 417 | 51.00 |
| Climbing Fish Inspired Sucker [130] | 110*80 | 26 | 3.00 |
| Clingfish Inspired Sucker [131] | Φ 25 | 45 | 15.00 |

Table 2: Parameter statistics on adhesion BUSRs.

Clingfish have suction cups on their oral sides, allowing them to adhere stably to rocks under tides and waves. Their suckers are almost passive and independent of muscle energy due to a special structure composed of pelvic and pectoral girdles [136]. They can withstand loads of 80 to 230 times their own weight. Inspired by clingfish, a passive sucker was developed (Fig. 5k). This research also discusses the impact of micropillars on the adhesion capacity of sucker lips [137].

Remora, a type of teleost fish, can adhere to the surfaces of boat hulls, sharks, rays, cetaceans, and sea turtles, and even divers. This ability is attributed to their specialized dorsal fins, which consist of soft disc lips and interior spinule-covered lamellae [132] (Fig. 5g). Adhesion and detachment are achieved through the rotating motion of the spinule-covered lamellae array [139][140].

The soft robotics laboratory at Beihang University has conducted extensive research on this topic. Initially, they fabricated a bio-inspired sucker with a pneumatic actuator (Fig. 5f), capable of withstanding loads of 340 times its weight [127]. Subsequently, untethered servo motor-driven suction BUSRs were developed (Fig. 5i). Furthermore, hydraulically driven suction cups have been integrated into aerial-aquatic robots (Fig. 5h), demonstrating excellent adhesion performance and even sensing capabilities [129][141].

Sliding adhesion allows rock climbing fish to traverse surfaces while enduring strong currents. Notably, climbing fish utilize their whole body as suction cups, capable of withstanding loads of 1000 times their weight [142][143]. However, the dynamic adhesion mechanism of climbing remains unclear. Some research has developed climbing fish-inspired BUSRs with multiple suckers to imitate dynamic adhesion (Fig. 5j). Ongoing studies are also exploring sliding adhesion principles in robots inspired by remora and even net-winged midge larvae [144]. The adhesive ability of an object or structure, such as a suction cup, is typically defined in terms of adhesion strength, which quantifies the adhesion force provided by the unit effective area. Adhesion BUSRs inspired by different biological mechanisms exhibit significant variations in size and shape, resulting in diverse adhesion performance (Table 2).

## 6. Environmental consideration of BUSRs

In addition to the requirements oriented toward functionality and bio-inspirations, the design of the BUSRs is substantially influenced by its intended operational environment. A variety of physical, chemical, and biological factors, including pressure, temperature, light, pH value and biodiversity, exert differing degrees of influence on both the design and the manufacturing processes of the BUSRs. These factors are often closely related to water depth [145][146][147]. Figure 6 illustrates the developmental trends of BUSRs, showing a marked increase in related research and a gradual expansion into deeper ocean regions in recent years. It also provides a visualization of how various oceanic factors influence BUSRs across different depths.

### 6.1 Ambient pressure

The increase in pressure associated with greater water depth is a critical factor influencing the design of BUSRs. In shallow water environments, such as near the water surface and laboratory water tanks with the depth of several tens of centimetres, pneumatic actuators remain a common choice for research at the initial stages [54][148][149].

The adaptability and high deformability of soft materials enable pneumatic actuators to achieve further advancements in actuation performance, safety and electronics-free fields [150]. Hydraulic actuators have advanced the development of BUSRs to greater depths. The hydraulic swimmers are capable of moving in the water from a depth close to the water surface in tank [40][151] to 18 meters in open sea [57], whereas the hydraulic grasping soft robots can operate normally at depths of several hundred meters [152], and in certain cases even up to 2,300 meters [153]. Nevertheless, in addition to the inconvenience caused by intricate drive and transmission components, under extremely high hydrostatic pressure, changes in the viscosity and frictional characteristics of the hydraulic fluid present a major obstacle to the ability of hydraulic BUSRs to operate at progressively greater depths [154]. Electrohydraulic actuators offer advantages over conventional hydraulic systems in deep-sea environments. Such actuators employing specially formulated dielectric liquids to transmit driving forces have enabled soft fish to reach depths of over 4,000 meters for untethered swimming [155].

Compared with complex hydraulic systems, DEA can be more readily optimized to withstand deep-sea hydrostatic pressure. Although high pressure acting on the DE membrane may reduce the attainable strain and even cause irreversible damage through stress concentration and high-voltage breakdown, it does not inherently eliminate the actuation capability. Moreover, performance optimization through copolymer modification has shown promising results [156], and DEA-driven soft robotic fish have successfully demonstrated operation in the Mariana Trench [41].

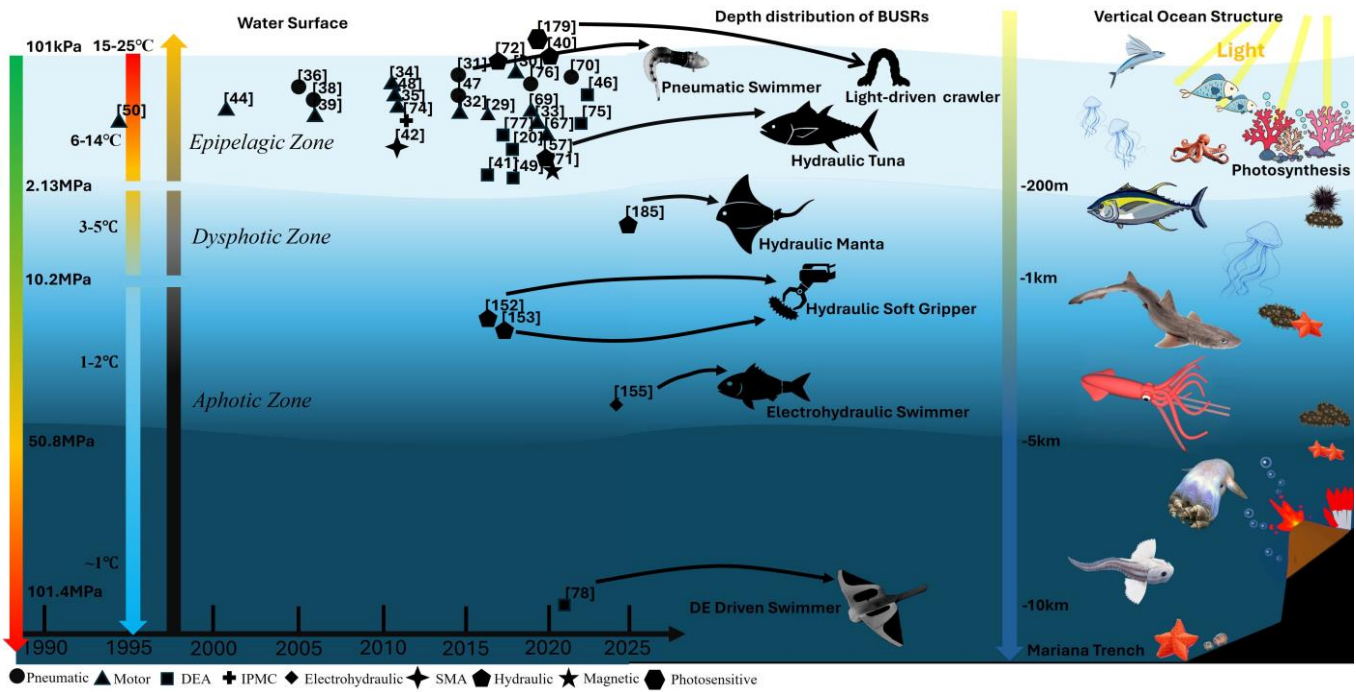

Fig. 6: The developmental trends of BUSRs in terms of both temporal and depth dimensions, as well as the vertical distribution within the marine ecological architecture and the depth ranges associated with different actuation modes of BUSRs.

To enable deep-sea exploration, the overall structural layout, material selection, energy systems, and packaging technologies of soft robots have been extensively optimized and refined. Decentralized design approach mitigates the shear stress imposed by extreme hydrostatic pressure on electronic components by minimizing rigid contacts and embedding them within soft materials such as silicone and hydrogel [41]. Furthermore, the integration of flexible circuits, liquid metal circuits, and gap-free capacitors respectively addresses the system's communication and power delivery requirements [157].

## 6.2 Temperature

Thermal variations represent a critical consideration in the design of underwater robots, and the near-freezing temperatures in the deep sea constitute a significant challenge [157][158]. The soft materials commonly employed in fabrication of soft robots not only stiffen at low temperatures, thereby hindering their mechanical motion, but also suffer reduced durability due to increased brittleness [159][160][161]. In particular, for heat-induced deformation materials such as thermal SMA, extremely low temperatures markedly diminish their response speed while increasing power consumption [162][163]. Although SMAs are highly sensitive to temperature, the intrinsic microstructure of the alloy renders them relatively insensitive to the high pressures encountered in the deep sea. A research inspired by deep-sea snails have suggested that SMA-driven underwater robots could operate in a high pressure environment equivalent to 3,000 meters deep [164].

Temperature significantly influences both the stiffness and dielectric constant of DE materials [157], and extremely low temperatures are associated with reduced strain capacity and slower response of DEAs [165][166][167]. The actuation performance of DEAs in deep-sea environments can be improved by lowering the glass transition temperature [156] or by increasing the dielectric constant of the dielectric elastomer [168].

Also, as an electrically driven actuator, although studies have demonstrated that IPMCs remain operable under extremely low-temperature conditions, even down to −140 °C [169]. The performance of IPMC is strongly influenced by low temperature. Similar to DEAs, IPMCs must contend with the potential damage, cracking caused by changes in material properties. In addition, their actuation relies primarily on ion migration, and under low-temperature conditions, the electrolyte is prone to freezing, ion mobility becomes severely restricted because of the decline in hydration capacity and leading to degraded performance or even complete loss of functionality [170][171]. For the protection and sustainable development of the marine environment, numerous studies have explored the use of biodegradable materials to replace conventional silicone or polyurethane in underwater soft robots, including gelatine, chitosan, and hydrogels [172][173][174]. Temperature exerts a significant influence on both hydrolytic and enzymatic degradation of these materials. Low temperatures slow down the chemical reactions underlying hydrolysis and markedly reduce the activity of natural enzymes, such as proteases and glucanases, thereby impeding enzymatic degradation as well [175][176]. As a result, under low-temperature conditions, particularly in deep sea, the decomposition efficiency of biodegradable soft robots is severely limited.

*6.3 Light and biodiversity*

Light serves as the primary energy source for the vast majority of life, and the underwater light environment is closely linked to patterns of marine biodiversity [177][178]. Both factors exert significant influence on the design considerations of BUSRs.

As photosensitive materials, both liquid crystal elastomers (LCEs) and liquid crystal networks (LCNs) can function as actuators for soft robots, enabling behaviours such as crawling on the water surface, swimming, or jumping underwater [179][180]. In addition, thermal Marangoni effect enables photothermal actuators to function as effective drivers for underwater soft robots [181]. The intensity of light across different wavelengths attenuates progressively with ocean depth and eventually becomes negligible. Therefore, regions below approximately 1,000 meters are generally designated as aphotic zones. Consequently, light-driven BUSRs are generally limited to operating on the water surface or within shallow zone.

Light also plays a critical role in shaping the environmental perception capabilities of underwater soft robots. In shallow waters, where visible light is abundant, most soft robots rely on vision-based methods, such as cameras, to acquire environmental data. In contrast, at greater depths where visible light is absent, contact sensors serve as the direct means of environmental perception [187], while infrared transceivers [41] and laser sensors [155] function as non-contact alternatives to cameras.

Biodiversity constitutes a key characteristic of marine ecosystems and is shaped by multiple factors, including hydrostatic pressure, temperature, ocean currents, crustal movements, and anthropogenic activities [188][189][190]. Different aquatic organisms occupy distinct depth zones within the ocean, serving not only as sources of inspiration for bionic design but also as potential interaction counterparts for BUSRs. For instance, soft robotic grippers developed for aquatic products harvesting or biological sampling must account for the economic or research value of the target species, ensuring capture is conducted as intact, efficient, and safe as possible. This imposes stringent requirements on the design of end-effectors and the choice of compliant materials [186][191]. Conversely, minimizing interactions with marine organisms, or saying, remaining undetected by them, also represents a crucial mode of interaction. In this context, underwater soft robots constructed from transparent materials can effectively camouflage themselves within the marine environment, thereby enabling closer and more prolonged observation of marine life without disturbance [75][192].

From another perspective, biological activities also shape the ocean chemical environment, whether through the photosynthesis of algae or the metabolic processes of marine mammals. In addition to the influence of atmospheric carbon dioxide and crustal activity at the seabed, the pH of seawater is significantly regulated by life processes [192]. In surface waters, intense photosynthetic activity tends to increase alkalinity. For mesopelagic and deep-sea, although local acidification may occur in certain seabed regions or nutrient-rich areas, causing dynamic variations in the distribution of ocean acidity and alkalinity [193][194], the pH generally decreases and approaches neutrality. Commonly used materials in soft robotics, such as silicone and polyurethane, are largely insensitive to weakly alkaline conditions and exhibit good chemical stability, making them well suited for deep-sea applications. In contrast, natural biodegradable materials, including gelatine and chitosan, degrade slowly under the combined conditions of low temperature and weak alkalinity [195][196], which limits the operational depth of biodegradable soft robots.

## 7. Filed Application of BUSRs

Currently, the vast majority of BUSRs investigated in this survey remain in the experimental phase. This indicates that the development of BUSRs is still in its early stages. Presently, only a few practical application cases have been achieved, primarily through BCF and MPF BUSRs. The research on the morphology and motion principles of these two types began earlier, which led to the relevant bionic technologies.

To date, the field applications of BUSRs in the civilian sector include ocean exploration, detecting and tracking water surface objects [197], underwater archaeology [182], and monitoring ocean organisms [57]. Furthermore, several technology companies, in collaboration with universities, research institutes, governmental bodies, and other research and functional organizations, have developed a variety of products. Some of these have been specifically designed for the consumer market and have already been officially released to the public, while customized solution packages are also offered to clients. Such as the manta and penguin inspired swimmers, from EvoLogics Ltd. The former belongs to the BOSS project, which has been supported and developed under the auspices of the German Federal Ministry for Economic Affairs and Energy [184]. The latter represents a mass-market consumer product, with a length of 1.12 m, a maximum cross-sectional diameter of 0.3 m, a weight of 25 kg, a payload capacity of 3 kg, a cruising speed of up to 5 m/s, and a single-operation endurance of up to 10 hours. A bionic shark robot developed by Boya Gongdao, Ltd, China, has also been unveiled at exhibitions. The BS-500 measures 1.85 m in length, weighs 500 kg, and is capable of operating at depths of up to 300 m [185]. In addition, the company has also developed other bionic products, such as an arowana and a manta ray.

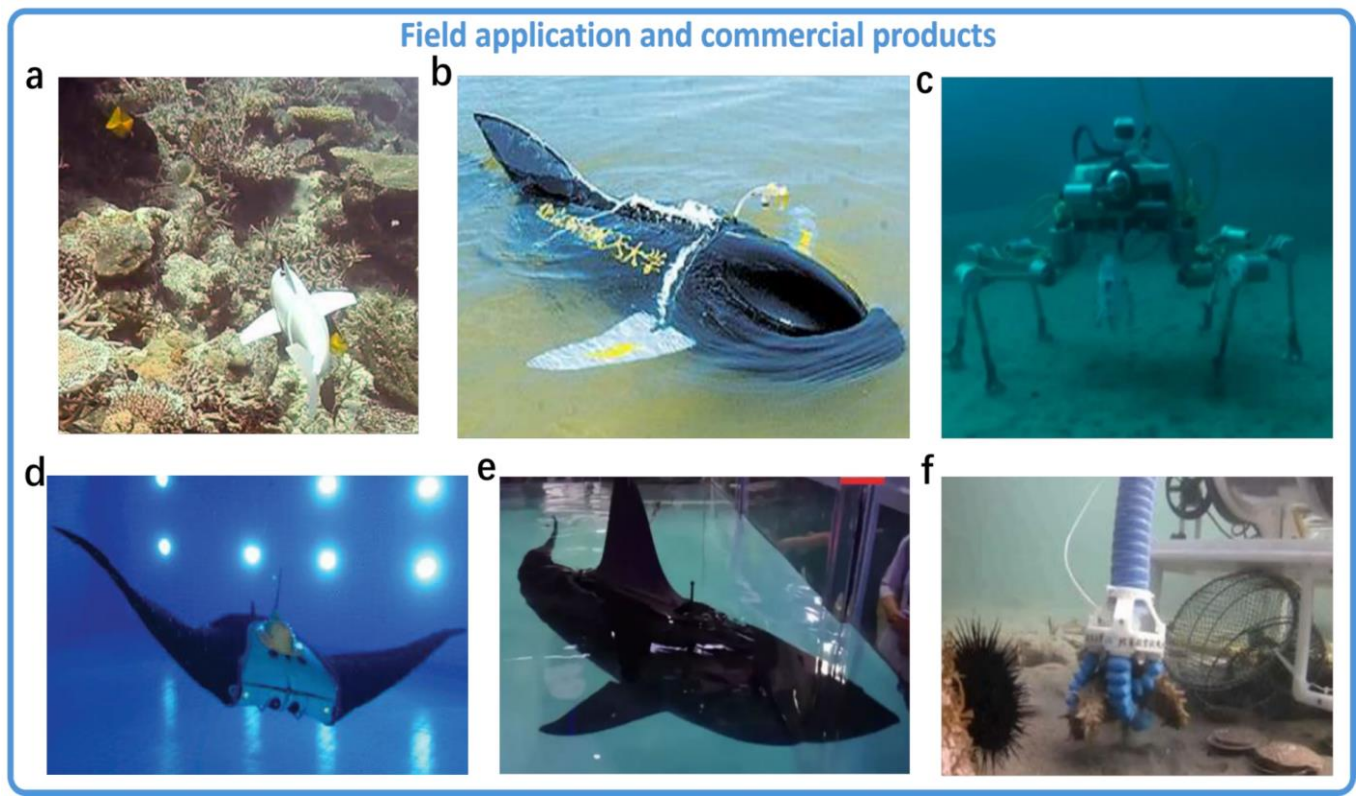

Fig. 7: Field application and future solutions. (**a**) Hydraulic tuna-inspired USR for detecting marine animals. Reproduced with permission from [57], copyright 2018, AAAS. (**b**) BCF USR for underwater archaeology. Reproduced with permission from [182], copyright 2005, Guangming Xie. (**c**) Legged ROV with soft gripper. Reproduced with permission under CC BY 4.0 license from [183]. (**d**) Biomimetic soft manta inspired robot, developed by Evologic Ltd, Germany. Reproduced with permission from [184], copyright 2022, IEEE. (**e**) Commercial biomimetic shark robot developed by Boya Gongdao, China [185]. Reproduced with permission from Boya Gongdao, Ltd. (**f**) Soft gripper for sea cucumber capture. Reproduced with permission from [186], copyright 2021, Sega publications.

The technological pathways of BUSRs resemble a combination of bioinspired principles and traditional underwater robots. The latter heavily rely on exteroceptive sensors for altitude control and navigation, such as sonar, infrared sensors, cameras, and Doppler velocimeters. As mentioned before, these sensors require rigid structures for containment and protection, which is against the compliance of soft robots and adds additional load. Moreover, the water pressure resistance capacity of rigid structures is a critical factor undermining their operational depth.

Additionally, soft grippers can be integrated with ROV to accomplish underwater tasks, including biological sampling and litter collection [183][186][198]. The relevant application cases are illustrated in (Fig. 7f). Compared to rigid grippers, soft grippers offer greater environmental suitability and underwater performance due to the compliance of soft materials and relatively simple driving schemes [199]. Image and distance sensors provide positional information about objects, which is essential for precise grasping, but achieving such high-intensity environmental awareness can be challenging for soft robot integration. Therefore, the collaboration between ROVs and soft grippers represents an ideal complementary operating mode at present.

## 8. Discussion

Over the past decade, impressive examples of bio-inspired robots designed for underwater environments have emerged. BUSRs is gradually moving from the laboratory stage to practical applications in open waters, and also showing a trend of deep-sea, specialized and intelligent development. However, questions remain regarding how to fully leverage the inherent advanced interactive features of BUSRs in complex underwater settings and translate these capabilities into practical engineering applications. By examining how aquatic creatures interactively perceive and operate within their surroundings, we envision more intelligent robots that can adapt seamlessly, to integrate with and benefit from underwater environments.

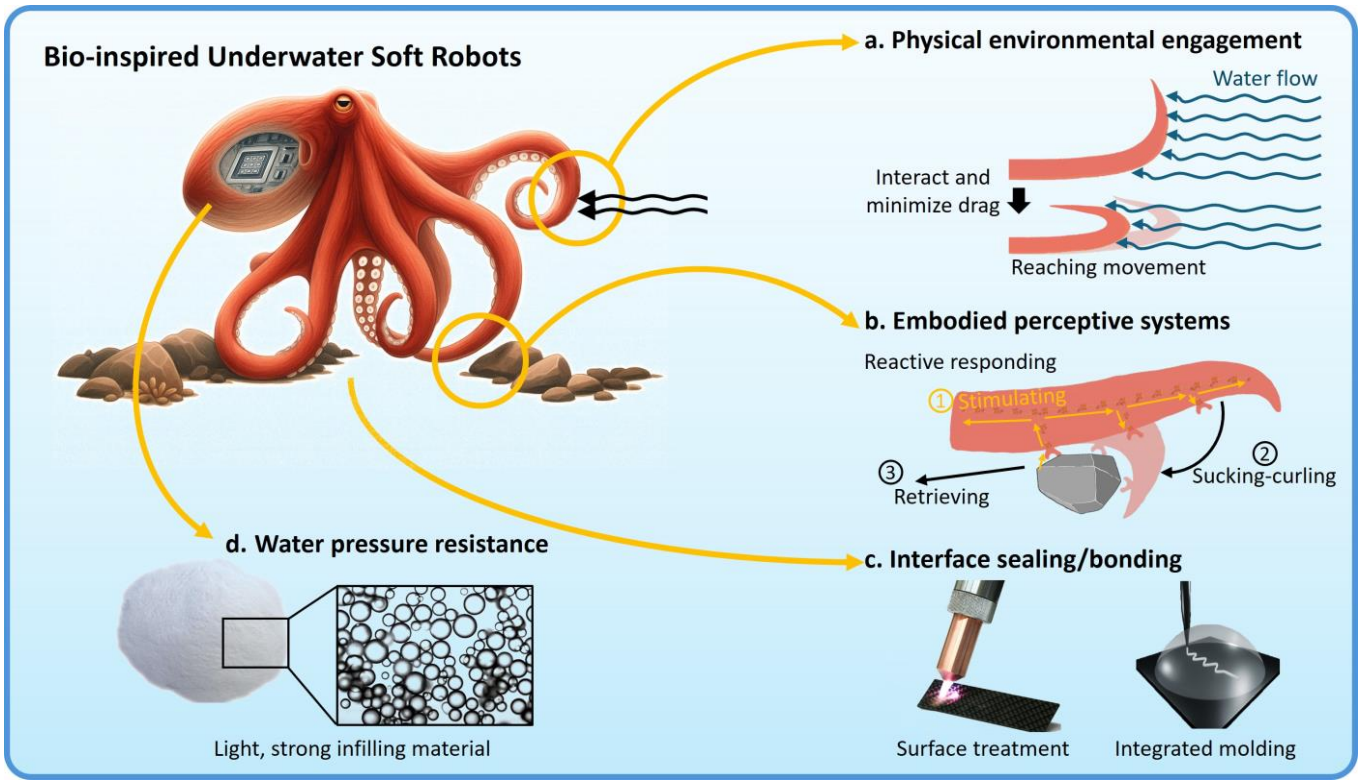

Fig. 8: Potential developments for future bio-inspired underwater soft robots. Embodied intelligence for effective environmental interactions could be considered in both (a) Physical environmental engagements and (b) Embodied perceptive system. Practical engineering solutions could be utilized to achieve better (c) Water pressure resistance and (d) Interface sealing/bonding of underwater soft robots.

Current challenges and potential future directions in this field primarily concern advancing the understanding and utilization of the relationship between interaction, proprioception, and embodied intelligence in BUSRs, alongside the general challenges of ocean engineering, such as design and manufacturing, material selection, and system integration.

*8.1 Embodied Intelligence*

Aquatic creatures utilize their embodied intelligence to interact effectively with their environments. The embodied intelligence for bio-inspired underwater soft robots should consider both of their physical environmental engagement (Fig. 8a) and embodied perceptive systems (Fig. 8b). Classic examples for efficient physical engagement include the passive interaction of soft-bodied dead fish with oncoming vortices to achieve propulsion [200] and the ability of an octopus arm to naturally form a typical reaching shape during underwater interactions (Fig. 8a), reducing drag and enabling efficient movement [12][14]. Similarly, remora fish exploit water flow dynamics to enhance adhesion strength [128]. These examples illustrate how aquatic creatures' lintrinsic intelligent mechanical structures engage with external conditions. Such adaptations significantly simplify control mechanisms and improve the manipulation efficiency of BUSRs in complex environments.

Embodied perceptive systems also play a crucial role in the environmental interaction of BUSRs. The ability to sense external conditions and respond quickly is essential for robotic intelligence when adapting to dynamic environments. For instance, the lateral line system in fish allows them to sense water flow and adjust their swimming accordingly [201]. Similarly, the vestibular system enables fish to rapidly modify their body posture for stability in turbulent wave conditions [202]. Octopuses, on the other hand, utilize their arm-sucker integrated neuronal network to detect target attachments, and reactively stimulate corresponding arm movements such as sucking-curling and retrieving, under the low-level computation embodied in the arm itself (Fig. 8b) [203][204][205]. While numerous underwater sensing components have been proposed in recent years [206][207], more integrated perceptive systems could advance higher-level perceptive models [208] and enable more seamless sensing-actuation coordination [209][210], enhancing BUSRs' adaptability to changing underwater environments.

*8.2 Potential Practical Engineering Solutions*

Although BUSRs demonstrate exciting possibilities for underwater tasks inspired by nature, transitioning them into practical applications presents significant challenges that require effective engineering solutions. Among these, pressure resistance is a critical factor for underwater systems, as it directly determines the operational depth of the equipment (Fig. 8d). Many underwater organisms utilize hydrostatic bones or muscles to balance internal and external pressures [211]. Given that water exhibits extremely high bulk modulus, early research drew inspiration from this by employing insulating liquids such as silicone oil to fill the electronic compartments of underwater robots. However, this method is prone to leakage issues and adds extra weight to BUSRs. Micro hollow ceramic or glass spheres [212] possess extraordinary pressure resistance, making them highly suitable for use in BUSRs. With excellent insulation proper ties, micro hollow ceramic, or glass spheres do not interfere with the normal operation of the circuit system. Additionally, their low density [213][214] helps counterbalance the weight of other components, effectively adjusting the overall equilibrium between weight and buoyancy.

Compared to pure soft robots, soft-rigid hybrid structures remain one of the most stable and feasible solution in the near term, given the current limitations of electronic and power systems. Consequently, excellent sealing performance, enabled by advanced assembly technologies, has become a critical factor (Fig. 8c). The combination of conventional rubber sealing rings with precision machined metal shells demonstrates excellent sealing effects at depths exceeding 10,000 meters underwater. However, this machine's technology is not applicable to soft materials because of their lower strength. Plasma surface treatment is a technique that has shown excellent results in assembling soft materials with various bonding partners [215][216], ensuring exceptionally uniform bonding at assembly seams and resulting in higher assembly accuracy to prevent uneven seams that could impact hydrodynamic characteristics, especially for small and micro BUSRs. Integrated molding techniques could also be employed to significantly minimize assembly procedures prone to errors and leaks in underwater soft robots. For example, Direct Ink Writing (DIW) 3D printing [217] combined with Freeform Reversible Embedding (FRE) technology can print structures within suspended hydrogels. The buoyancy provided by the surrounding hydrogel eliminates the need for additional support materials [218][219]. This method enables precise and efficient printing of complex geometries with a wider variety of soft materials [220], making it particularly well-suited for fabricating micro-scale or intricate soft robots.

## 9. Conclusion

This paper provides a precise definition of BUSR and offers a comprehensive, multi-faceted survey of existing prototypes worldwide. It serves as a valuable resource for novice researchers in the field, enabling a swift and thorough understanding of the development of bio-inspired underwater soft robots when considering different desired functions, bio-inspirations, ambient pressure, temperature, light, and biodiversity.

Additionally, the paper discusses the potential of embodied intelligence in BUSRs, highlighting the challenges associated with their applications and proposing several practical solutions to address these challenges. The analysis in this review offers original insights into the future direction of bio-inspired underwater soft robots.

## Acknowledgements

This work was supported by the Southern University of Science and Technology Founds for High-level platform construction (G030330002 and G030330003 to Z.X.) and National University of Singapore start-up grant RoboLife (A-0009125-02-00 to C.L.).

J Wang et al